\documentclass[pdflatex,sn-mathphys-num]{sn-jnl}

\usepackage{graphicx}%
\usepackage{csquotes}
\usepackage{subcaption}
\usepackage{listings}
\usepackage{color,soul}
\usepackage{multirow}
\usepackage[title]{appendix}%

\usepackage{natbib}

\begin{document}

\title{Topic Discovery and Classification for Responsible Generative AI Adaptation in Higher Education}

\author*[1]{\fnm{Diane} Myung-kyung \sur{Woodbridge}}\email{diane@studystudio.ai}
\author[2]{\fnm{Allyson} \sur{Seba}}\email{aseba@usc.edu}
\author[1]{\fnm{Freddie} \sur{Seba}}\email{freddie@studystudio.ai}
\author[1]{\fnm{Aydin} \sur{Schwartz}}\email{aydin@studystudio.ai}

\affil*[1]{\orgname{StudyStudio.ai}, \orgaddress{ \city{San Francisco}, \state{California}, \country{USA}}}

\affil[2]{\orgdiv{Computer Science}, \orgname{University of Southern California}, \orgaddress{\city{Los Angeles},  \state{California}, \country{USA}}}

\abstract{
As generative artificial intelligence (GenAI) becomes increasingly capable of delivering personalized learning experiences and real-time feedback, a growing number of students are incorporating these tools into their academic workflows. They use GenAI to clarify concepts, solve complex problems, and, in some cases, complete assignments by copying and pasting model-generated contents. While GenAI has the potential to enhance learning experience, it also raises concerns around misinformation, hallucinated outputs, and its potential to undermine critical thinking and problem-solving skills.

In response, many universities, colleges, departments, and instructors have begun to develop and adopt policies to guide responsible integration of GenAI into learning environments. However, these policies vary widely across institutions and contexts, and their evolving nature often leaves students uncertain about expectations and best practices.

To address this challenge, the authors designed and implemented an automated system for discovering and categorizing AI-related policies found in course syllabi and institutional policy websites. 

The system combines unsupervised topic modeling techniques to identify key policy themes with large language models (LLMs) to classify the level of GenAI allowance and other requirements in policy texts. The developed application achieved a coherence score of 0.73 for topic discovery. In addition, GPT-4.0-based classification of policy categories achieved precision between 0.92 and 0.97, and recall between 0.85 and 0.97 across eight identified topics.

By providing structured and interpretable policy information, this tool promotes the safe, equitable, and pedagogically aligned use of GenAI technologies in education. Furthermore, the system can be integrated into educational technology platforms to help students understand and comply with relevant guidelines.
}
\maketitle 

\section{Introduction}
Since the release of OpenAI's ChatGPT \cite{chatgpt_2024}, a generative artificial intelligence (GenAI) using a large language model (LLM), in November 2022, various companies, including Cohere \cite{command_2025}, Anthropic \cite{claude_2025}, Google \cite{gemini_2024}, and Meta \cite{llama2025}, have introduced powerful LLM models and their updates.  These models, and the applications built on them, are transforming many aspects of daily life and work. These models, and the applications built on them, are transforming many aspects of daily life and work—from customer service chatbots and software development to marketing with AI-generated media and even drug discovery \cite{sengar2024generative}. The use and applications of LLMs have largely impacted education and learning as well.  Thanks to their ability to generate personalized responses, deliver real-time feedback, and enhance efficiency, students have rapidly adopted tools powered by these models \cite{law2024application}.

Despite the benefits of GenAI in education, there have been many concerns including academic misconduct, hallucination, biases, legal and ethical use, privacy, and security \cite{chan2023comprehensive}. Crucially, without clear guidance and best practices, GenAI tools may hinder rather than help learning—potentially weakening students' problem-solving abilities, critical thinking, and foundational understanding \cite{qawqzeh2024exploring}.  Recent research shows that dependency on GenAI tools can cause memory loss, procrastination, and worsen academic performance \cite{abbas2024harmful}.

In education, higher education institutes first started discussing incorporating GenAI policy and guidelines into student honor codes and course syllabi. As GenAI is new to many, higher education institutes organized committees including experts in artificial intelligence, education, policy, ethics, and other disciplines to endeavor to provide templates, examples, and general guidelines \cite{dabis2024ai}.

As AI policies are established based on ethical considerations,  learning objectives, and pedagogical approaches of the discipline, instructor, course, and even an assignment level, they vary at the institution, department, course, lesson plan, and assignment levels \cite{ghimire2024guidelines}\cite{mcdonald2024generative}. 
This variability can result in vague or inconsistent guidance, contributing to confusion and inequity in student learning experiences. As such, clear articulation of expectations including allowed usage, attribution, data validation, and cautions on information release is critical \cite{assesment_and_evaluation_in_higher_education}.

Because GenAI models and tools evolve rapidly, AI-related policies and guidelines must undergo regular review and adaptation to keep pace with technological changes \cite{school_of_information_student_research_journal}.  This would require organizing committees even at a discipline level to understand both risks and benefits of adapting the up-to-date GenAI models and applications, and incorporate them into curricula \cite{hct}. 

Therefore, establishing effective AI policies in education demands significant resources, including domain expertise in AI systems and their pedagogical implications across diverse subject areas, as well as time commitments for iterative review and refinement \cite{nikolic2024systematic}.
These requirements can pose substantial challenges for smaller or under-resourced institutions.
To accommodate this, identifying recurring themes and categories in existing policies can offer a useful starting point to lay the groundwork for scalable and adaptable frameworks. Such frameworks can help GenAI integration in under-resourced higher education and eventually support policy development in K–12 environments, where adoption is currently lagging \cite{ghimire2024guidelines}.

In this paper, the authors present a scalable data pipeline for collecting and analyzing up-to-date GenAI policies and guidelines from educational institutions. We apply topic discovery algorithms to uncover key themes with high coherence, which can inform the design and review of future policies. Additionally, we evaluate the use of LLMs to classify categorical values within policy documents, which can be used to provide a concise guideline and/or applied to educational technology systems utilizing GenAI to limit its functionalities and comply with policies.

Section~\ref{sec:system_design} outlines the system workflow, encompassing data collection, topic discovery, and category classification. Section~\ref{sec:results} presents the collected data along with the results from topic discovery and category classification. Finally, Section~\ref{sec:discussion} provides the conclusion and discussion of this study.
\section{System Design}
\label{sec:system_design}

\begin{figure}
    \centering
    \includegraphics[width=0.9\linewidth]{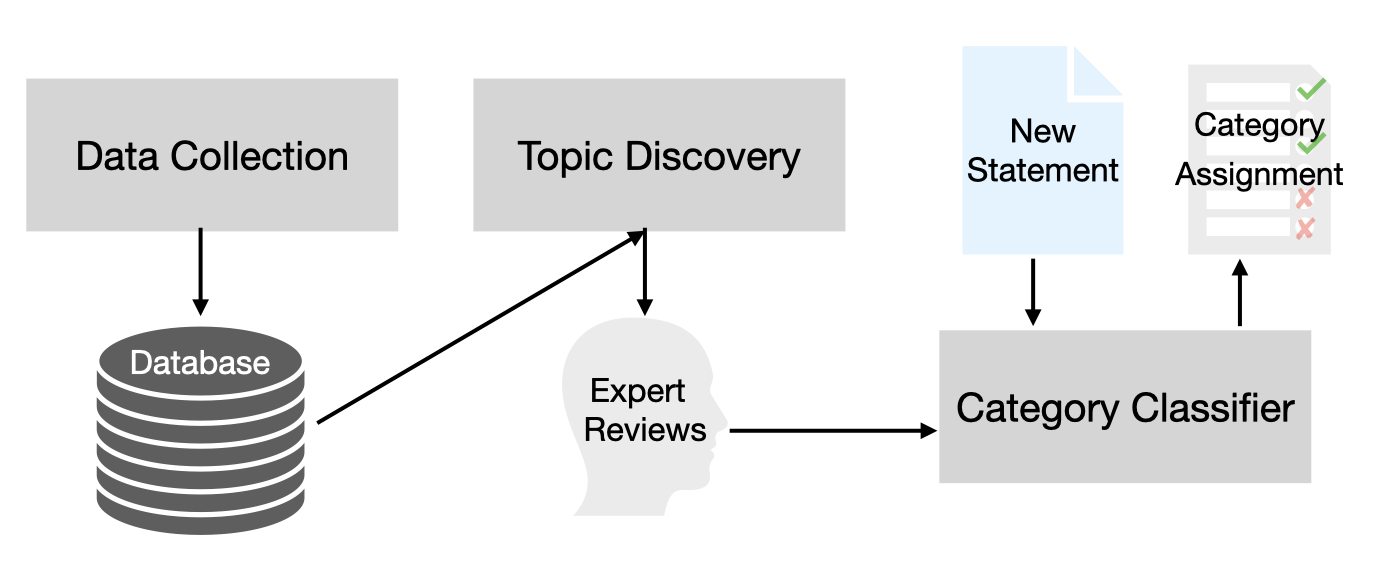}
    \caption{System Workflow: Collected data from the data collection pipeline is used for topic discovery. Outputs with high scores are reviewed by experts and used in the category classification step. }
    \label{fig:system_overview}
\end{figure}

In order to develop a data-driven topic discovery and classification system, the authors designed a workflow to continuously search and collect GenAI policies in accredited post-secondary education institutions. The system periodically performs topic discovery with newly updated data collection to find relevant topic categories and evolve the available categories and corresponding options. As the topic discovery step returns a list of candidate topics that are hard to interpret in nature, human experts review and translate them into human-readable formats. When a user provides a policy statement as input, the category classifier returns a list of mentioned categories and their corresponding values (See Figure~\ref{fig:system_overview}).

\subsection{Data Collection Pipeline}
\label{sec:data_collection}
Since the launch of OpenAI's ChatGPT \cite{chatgpt_2024} in November 2022, the use of GenAI in educational settings has increased dramatically. A recent survey shows that 89\% of college students have used ChatGPT and other GenAI tools for their school work, including assignments, quizzes, tests, and outlining a paper \cite{westfall_2023}. However, with concerns including plagiarism, hallucination, copyrights, and sensitive information release to a model, many higher education institutes have started to come up with university-wide committees and put efforts into providing guidelines in GenAI uses in various academic settings including course works and research, and expect to tailor to each class or subtask levels.

Given that these efforts majorly started in a few higher education institutions first, and still, many discussions and debates are going on, many institutions do not have policies or have continuous changes in their statements and examples. To incorporate the most recent data and its trends, the authors built a continuous and scalable data collection pipeline. The collected data will be used for topic discovery and category classification in the next steps, where outputted candidate policy categories and underlying models could be periodically updated.

Firstly, we gained access to the U.S. Department of Education's database of accredited post-secondary institutions and programs (DAPIP, \cite{dapip}) to retrieve information about accredited institutions, including their names and information. 
The developed data pipeline performs web searches for GenAI policies of the institution, colleges/schools, departments, and courses under the institution using the institution's name. This scraped information in HTML or PDF is parsed, stored, and updated in MongoDB \cite{mongodb}, a document database, preserving its nested and hierarchical structure of the original data (See Listing~\ref{json_example}). 

\lstset{frame=tb,
  language=Java,
  aboveskip=3mm,
  belowskip=3mm,
  showstringspaces=false,
  columns=flexible,
  basicstyle={\small\ttfamily},
  numbers=none,
  breaklines=true,
  breakatwhitespace=true,
  tabsize=3
}

\begin{lstlisting}[caption= Example Data in JavaScript Object Notation (JSON), label=json_example]
{ "institutions": [
        {
            "_id": "univ_a",
            "name": "University A",
            "url": "https://univ_a.edu/url",
            "last_update": "2025-01-01 15:30:00",
            "colleges": [
                {
                    "_id": "univ_a_college_arts_sciences",
                    "name": "College of Arts and Sciences",
                    "url": "https://univ_a.edu/arts_sciences/policy",
                    "last_update": "2025-01-01 15:30:00",
                    "departments": [
                        {
                            "_id": "univ_a_college_arts_sciences_dept_physics",
                            "name": "Department of Physics",
                            "url": "https://univ_a.edu/arts_sciences/physics/policy",
                            "last_update": "2025-01-01 15:30:00",
                            "policy":[
                            {"timestamp": "2024-09-01 15:00:00", "text": "Students are permitted to use generative AI tools."},
                            {"timestamp": "2025-01-01 15:30:00", "text": "Students are permitted to use generative AI tools if they disclose any use of AI-generated material."}
                            ]
                        },
                        {
                            "_id": "univ_a_college_arts_sciences_dept_chemistry",
                            "name": "Department of Chemistry",
                            "url": "https://univ_a.edu/arts_sciences/chemistry/policy",
                            "last_update": "2024-09-01 15:00:00",
                            "policy":[{"timestamp": "2024-09-01 15:00:00", "text": "Students are permitted to use generative AI tools."}]
                        }
                    ],
                    "policy":[
                    {"timestamp": "2024-09-01 15:00:00", "text": "Students are permitted to use generative AI tools."}
                    ]
                },
                {
                    "_id": "univy_a_college_business",
                    "name": "College of Management",
                    "url": "https://univ_a.edu/management/url",
                    "last_update": "2024-12-24 15:30:00",
                    "departments": [
                        {
                            "_id": "univ_a_college_management_dept_accounting",
                            "name": "Department of Accounting", "url": "https://univ_a.edu/management/accounting/policy",
                            "last_update": "2024-09-01 00:00",
                            "policy":[
                            {"timestamp": "2024-09-01 00:00:00", "text": "Students are not permitted to use generative AI tools for assignments and assessments."}
                            ]
                        }],
                    "policy":["timestamp": {"2024-08-15 00:00:00", "text": "Students are not permitted to use generative AI tools for assignments."}]
                }
            ],
            "policy":[{"timestamp": "2024-08-15 00:00:00", "text": "Students may use generative AI tools only if their instructor allows."}]
        }
] }
\end{lstlisting}

This data collection pipeline runs periodically to include up-to-date policy documents that are used in the next steps.

\subsection{Topic Discovery}
\label{sec:topic_discovery_algorithm}

\begin{figure}[h!]
    \centering
    \includegraphics[width=0.9\linewidth]{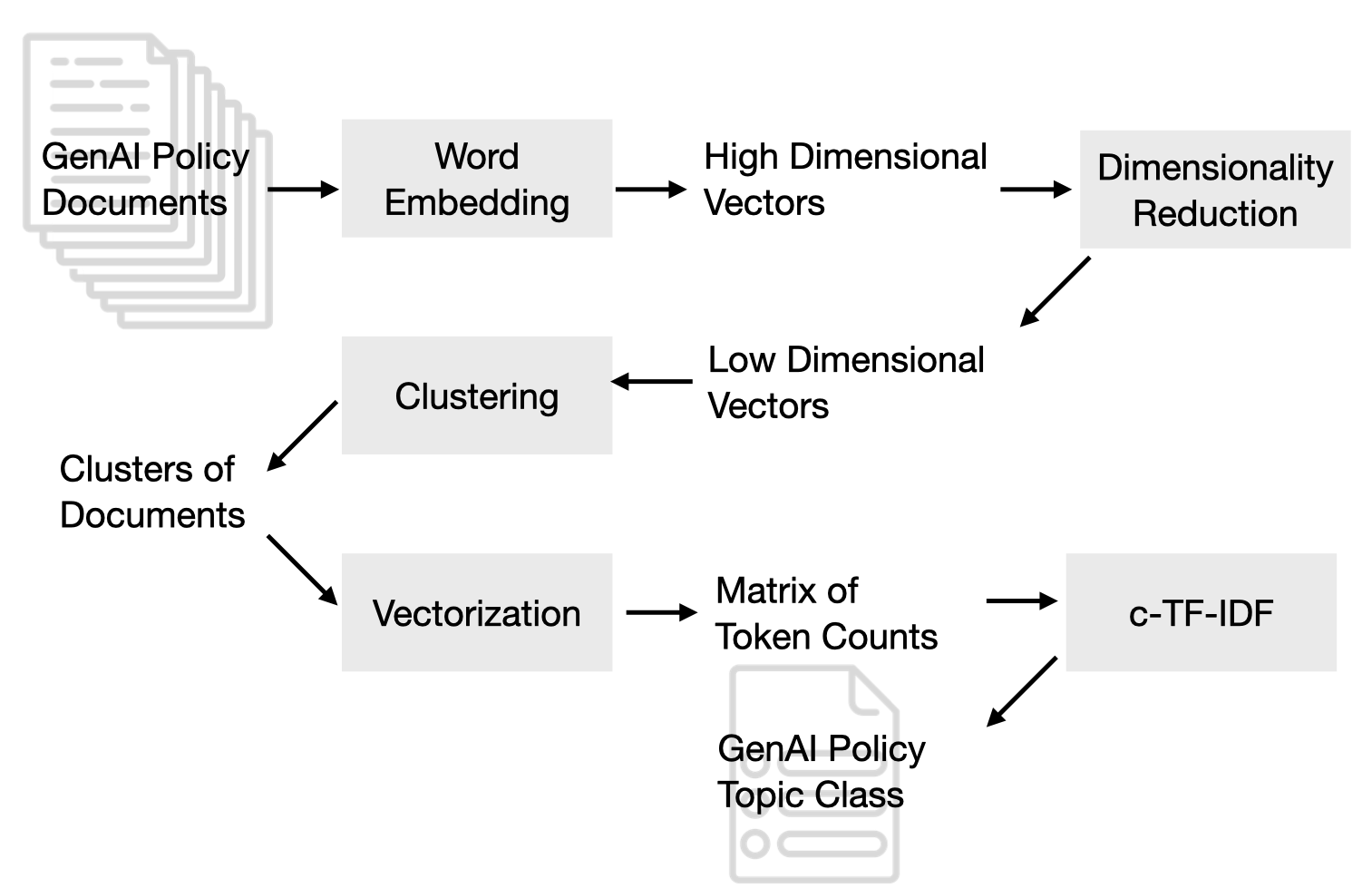}
    \caption{Workflow of Topic Discovery. (Please note that vectorization can also happen after the training.)}
    \label{fig:topic_discovery}
\end{figure}

In order to identify topics in AI Policy documents, the authors applied BERTopic with pre-trained transformer-based language models to generate embedding vectors, then clustered the embeddings, and applied a class-based term frequency-inverse document frequency (c-TF-IDF, $W_{x,c}$ in Equation~\ref{eq:c-tf-idf}) to generate topic representation \cite{grootendorst2022bertopic}. 

Transformers generate vector embeddings for words in the input document and capture their semantic relationships.  As this requires a huge amount of data and a high computing power to train neural network models, pre-trained transformer models are often used as a starting point. Examples of the pre-trained transformers in large language models include OpenAI's Generative Pretrained Transformer (GPT) \cite{brown2020language}, Cohere's Embed \cite{cohere_embed}, Bidirectional Encoder Representations from Transformers (BERT) \cite {alaparthi2020bidirectional}, and its variations, etc.

c-TF-IDF measures the importance of a word within a specific class or topic ($W_{x,c}$ in (\ref{eq:c-tf-idf})), using term-frequency(TF) and inverse document frequency (IDF) to highlight terms that distinguish one class from others.
\\
\begin{equation}
\label{eq:c-tf-idf}
W_{x,c}  =  \Vert  tf_{x,c} \Vert \times \log{(1+\frac{A}{f_x})}
\end{equation}

\;

\noindent where $tf_{x,c} $ and $f_x$ are the frequency of word $x$ in class $c$ and the frequency of $x$ across all classes accordingly. $A$ denotes the average number of words per class.

To achieve high-quality topic discovery using transformers and c-TF-IDF, the authors compared various transformer models and explored the impact of different pre-processing or fine-tuning processes, including dimensionality reduction, clustering, and vectorization.

To evaluate the performance of each step, the authors employed the coherence score, $Coh_v$, which quantifies how well words in an output set support each other. This is a key metric to assess the quality of output and its interpretability by humans \cite{roder2015exploring}.
\\

\begin{equation}
\label{eq:coherence_score}
    Coh_v = (P_{sw}, S^{one}_{set},\tilde{m}_{cos},  \sigma_a)
\end{equation}

\;

\noindent where $P_{sw}$ is the probability estimated using a boolean sliding window, which iterates over the document one word at a time, capturing the likelihood of a word appearing within the defined window.
$S^{one}_{set}$ represents a segmentation of the word set, which compares the words in the set to the total word set of the document.
$\tilde{m}_{cos}$ is an indirect confirmation measure, computed using cosine vector similarity to quantify the agreement between word vectors. Finally, $\tilde{m}_{cos}$ values from all pairs in $S^{one}_{set}$ are aggregated into a single coherence score using arithmetic mean, $\sigma_a$.

\subsubsection{Dimensionality Reduction}
As pre-trained transformer models result in high dimensional vectors which can make finding patterns and relationships harder, we applied Uniform Manifold Approximation and Projection  (UMAP, \cite{mcinnes2020umapuniformmanifoldapproximation}) for dimension reduction with different hyperparameters. 
UMAP seeks a low-dimensional representation of the data that maintains the relationships captured in the fuzzy topological structure by minimizing cross-entropy between the original high-dimensional and low-dimensional representations, (\ref{eq:entropy}).
\\
\begin{equation}
\label{eq:entropy}
    \sum_{e \in E} w_h(e) \log{\frac{w_h(e)}{w_l(e)}} +  (1- w_h(e)) \log{\frac{w_h(e)}{w_l(e)}} 
\end{equation}

\;

\noindent where $E$ is all possible 1-simplices, and $w_h(e)$ and $w_l(e)$ are weight of $e$ in the high and low dimensional cases. In UMAP, weights indicate a probability that the edge exists in the given dimensions and it aims to find a low-dimensional representation where the distribution of edge weights closely corresponds to the distribution in its high-dimensional representation. 


\subsubsection{Clustering}
To capture key topics in documents, we need to group the transformed low-dimensional vectors from UMAP. Clustering algorithms use unsupervised learning to group data so that items within the same cluster are more similar to each other than to those in different clusters.

In this study, the authors applied and compared K-means and Hierarchical Density-Based Spatial Clustering of Applications with Noise (HDBSCAN) with hyperparameter tuning. K-means begins by randomly selecting $K$ centroids for each cluster. The algorithm then assigns data points to the nearest centroid and recalculates the centroids. This process repeats until it reaches a minimum value for the sum of distances between data points and their centroids \cite{macqueen1967some}.
HDBSCAN is a density-based hierarchical clustering algorithm that transforms data points into a minimum-spanning tree, a subset of edges in a connected, undirected, and weighted graph that connects all vertices without cycles, and with the minimum total edge weight. HDBSCAN uses distance and minimum cluster size to create a hierarchical structure. The algorithm chooses clusters that exceed the minimum size limit and maximizes the stability score, $Stab(C_i)$, for a cluster, $C_i$ \cite{campello2013density}.
\\
\begin{equation}
\label{eq:hdbscan_stability}
    Stab(C_i) = \sum_{p_j \in C_i}(\lambda_{max}(p_j, C_i) - \lambda_{min}(C_i))
\end{equation}

\;

\noindent where $\lambda_{min}(C_i)$ is the minimum density level that $C_i$ exists and $\lambda_{max}(p_j, C_i)$ is the density level that $p_j$ no longer exists within $C_i$.

\subsubsection{Vectorization}
To represent document topics in a cluster, applying a count vectorizer with c-TF-IDF can be effective, where the count vectorizer converts text into a sparse vector with word frequencies. Since frequency may be relevant to the representative topic, a threshold value, minimum term frequency, is applied. This vectorization can occur before or after training a topic model. Applying it before training reduces the c-TF-IDF matrix size while applying it after training fine-tunes the representation.

\subsection{Category Classification}
After topic discovery in Section~\ref{sec:topic_discovery_algorithm}, experts in higher education review outputs with high coherence scores ($Coh_v$) and translate them into distinctive categories that humans can understand. 

Example outputs before the expert interpretation from the topic models are the followings. Once experts analyze topic model representations with corresponding representative policy documents, a list of discrete categories and possible values for each category are defined.

\begin{lstlisting}[caption=Example outputs from the topic discovery step, label=category_output]
    [ `unauthorized', `exam', `assistance', `assignment', `statement', `permission', `instructor', `clear', `use', `intelligence' ]
    [ `sensitive', `information', `data', `public', `confidential', `platforms', `business', `input', `mindful', `internal' ]
    [ `inaccurate', `material', `accuracy', `content', `responsible', `publication', `misleading', `biased', `hallucinations', `information' ]
\end{lstlisting}

For classification, the system takes the GenAI policy texts which may or may not include all the categories found in the topic discovery step into pre-trained large language models (LLMs) to return a value of each category. For instance, an input text may include information about categories such as ``allowed to use in learning", ``allowed to use for assignments", and/or ``allowed to use for assessment", and could be ``true" or ``false" if they appear.

To compare various LLMs, we utilized LangChain, an open-source platform that allows to create and connect LLMs  \cite{langchain} for benchmarking with selected data sets to find out which model works the best for classifying GenAI policies in the context of education and academia. Note that classification models should be regularly evaluated and updated as they tend to show different outcomes day-to-day, and new and more powerful models come out often.
For evaluating LLMs, we used precision and recall as a metric.

\begin{equation} Precision = \frac{TP}{TP+FP}\end{equation}
\begin{equation} Recall = \frac{TP}{TP+FN}\end{equation}

\;

\noindent where $TP$, $TN$, $FP$, and $FN$ are true positive, true negative, false positive, and false negative accordingly.
\section{Experimental Results}
\label{sec:results}
For discovering major topics in GenAI policies and benchmarking the performance of LLMs to detect and classify policy texts to topics, we collected data and fed them into the topic discovery step with different configurations. We selected a model and its outcome with the highest coherence score, which experts in higher education reviewed and converted outputs to a category that humans can understand. For category classification, randomly selected GenAI policy texts with corresponding categories and labels are inputted to different LLMs to compare which model outperforms.

\subsection{Data}
The data acquisition component of our developed pipeline involved scraping publicly available policies, guidelines, and templates related to the use of GenAI. These documents were sourced from a variety of universities and included both general guidelines and specific course syllabi sections.

Given the presence of templates detailing both permissible and prohibited uses of GenAI tools, we segmented the documents into individual paragraphs or sections. This segmentation facilitated more effective topic modeling and text classification, resulting in a total of 212 unique documents collected from 22 institutions.

To validate the performance of our text classification models, we received the help of experts, current faculty members with extensive experience serving on university-wide committees, who manually labeled the data, which was used as ground truth.

\subsection{Results}
\subsubsection{Topic Discovery}
\label{sec:topic_discovery}
In order to find the best topic discovery model for GenAI policies, the authors tested six different word embeddings with various algorithms and configurations at each step of dimensionality reduction, clustering, and vectorization. The six embedding models include Bidirectional Encoder Representations from Transformers (BERT) \cite{alaparthi2020bidirectional}, Cohere's Large Representation model \cite{cohere_embed}, and OpenAPI's Small, ADA, GPT-3.5,  and GPT-4.0 \cite{openai_vector_embeddings} \cite{openai_platform_2025}.

For each step, the authors chose a hyperparameter and increased its value to find the local maxima. In addition, if the pre-processing step itself degrades the outcome, the step is ignored for further steps.
For dimensionality reduction, coherence scores were measured with varying numbers of neighboring points ($n\_neighbors$) UMAP to consider.
For clustering, both K-means and HDBSCAN were compared. For HDBSCAN, we incremented $min\_cluster\_size$, the minimum number of data points within a cluster, to measure the coherence score. For K-means, we incremented $n\_clusters$, the number of clusters that we would allow, and evaluated its coherence score.
Vectorization was applied before and after training the topic model, with varying $min\_df$, a minimum value of a word to appear to be added to the representation. During the c-TF-IDF step, we compared a BM-25(Best Matching 25) weighting, which accounts for term frequency saturation and document length \cite{robertson2009probabilistic}, with a default weighting method, $W_{x,c}$ mentioned in Section \ref{sec:topic_discovery_algorithm}.

\begin{figure}
    \centering
    \begin{subfigure}[]{0.8\textwidth}
        \includegraphics[width=\linewidth]{"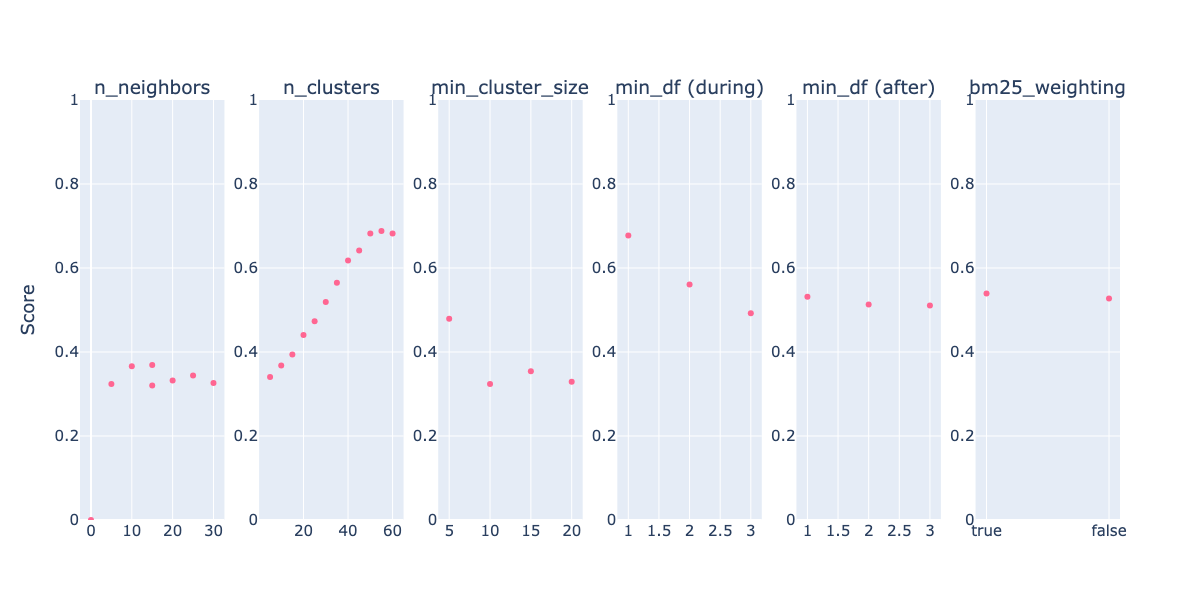"}
        \caption{Coherence Score of BERT}
    \end{subfigure}
    \begin{subfigure}[]{0.8\textwidth}
        \includegraphics[width=\linewidth]{"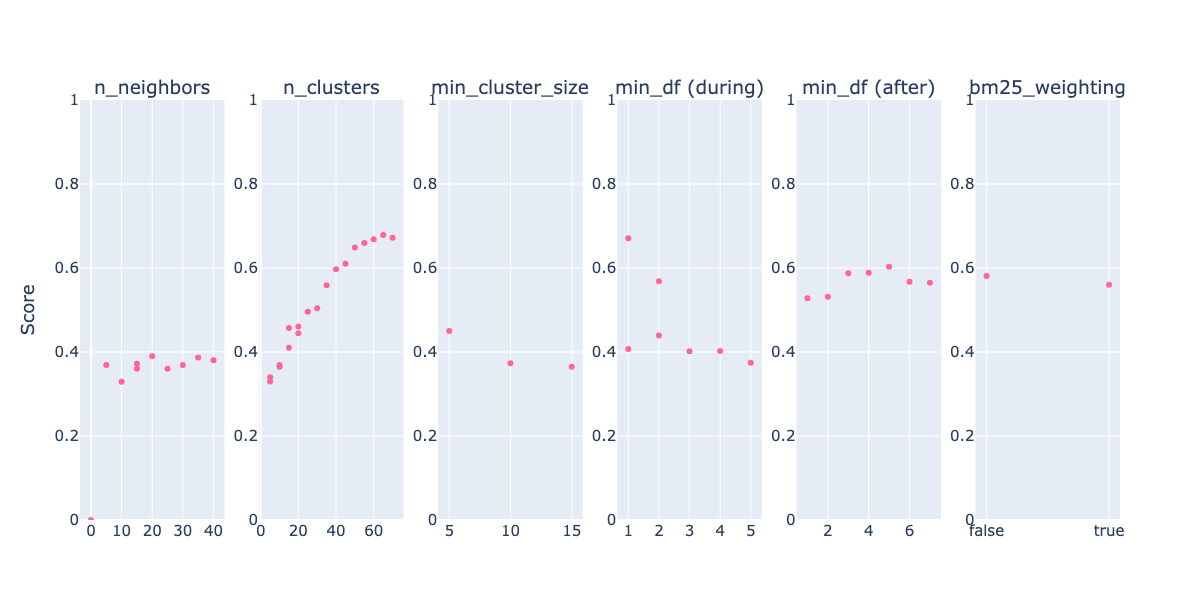"}
        \caption{Coherence Score of Cohere Large Representation}
    \end{subfigure}

    \begin{subfigure}[]{0.8\textwidth}
        \includegraphics[width=\linewidth]{"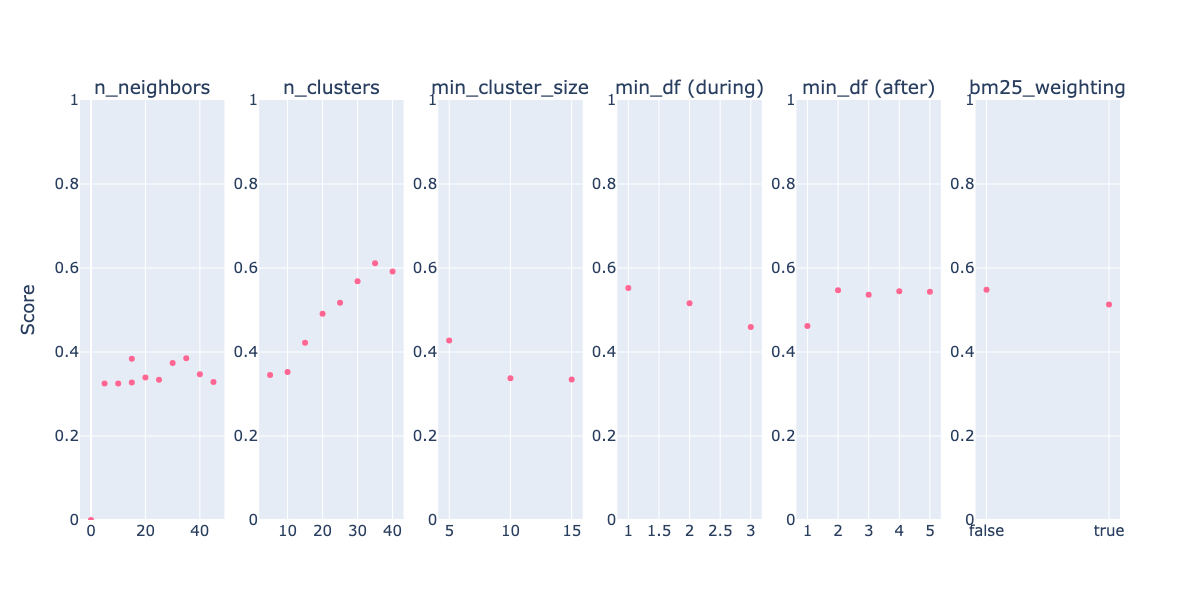"}
        \caption{Coherence Score of OpenAI Ada}
    \end{subfigure}    
    \caption{Coherence Score with Different Parameter Types and Values using Various Embeddings - BERT, Cohere Large Representation, and OpenAI Ada }
    \label{fig:parameters_score_1}
\end{figure}

\begin{figure}
    \centering
    \begin{subfigure}[]{0.8\textwidth}
        \includegraphics[width=\linewidth]{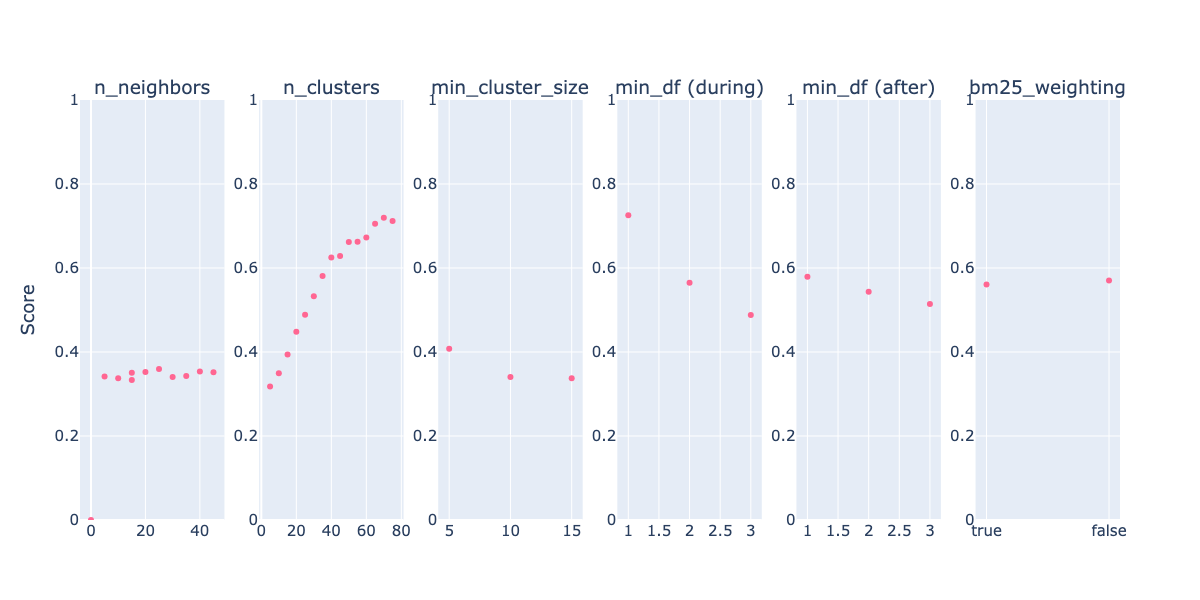}
        \caption{Coherence Score of OpenAI Small}
    \end{subfigure}
    \begin{subfigure}[]{0.8\textwidth}
        \includegraphics[width=\linewidth]{"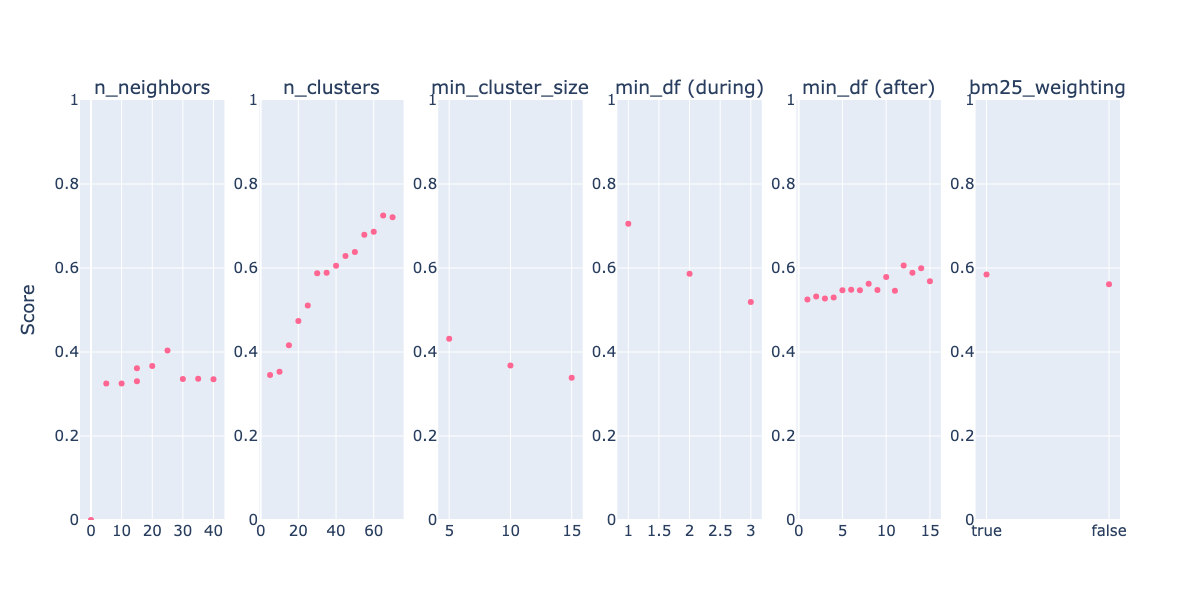"}
        \caption{Coherence Score of OpenAI GPT-3.5}
    \end{subfigure}

    \begin{subfigure}[]{0.8\textwidth}
        \includegraphics[width=\linewidth]{"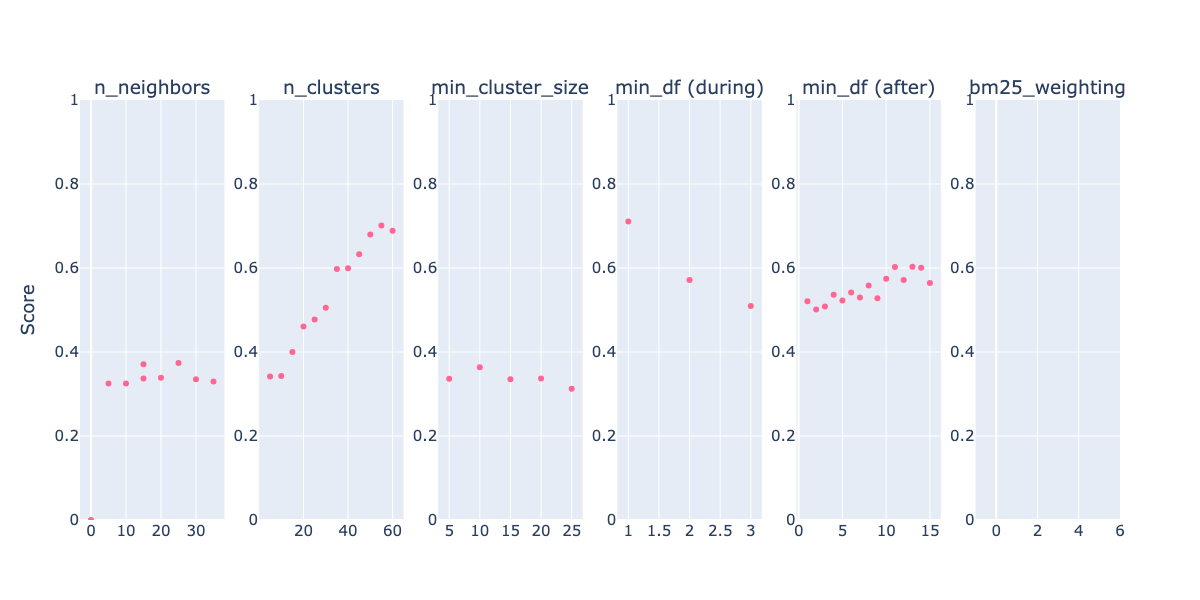"}
        \caption{Coherence Score of OpenAI GPT-4.0}
    \end{subfigure}    
    \caption{Coherence Score with Different Parameter Types and Values using Various Embeddings - OpenAI Ada, Small, GPT-3.5, and GPT 4.0 }
    \label{fig:parameters_score_2}
\end{figure}

\begin{figure}
    \centering
    \includegraphics[width=\linewidth]{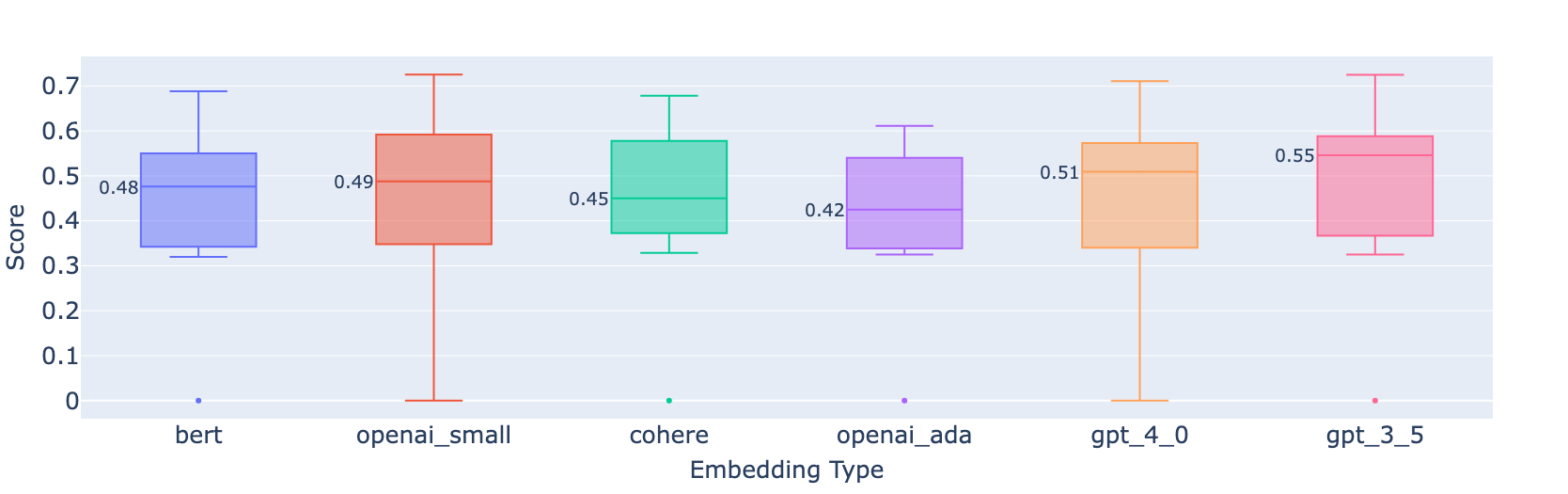}
    \caption{Embedding Types and Coherence Scores}
    \label{fig:embedding_types_results}
\end{figure}

Figure~\ref{fig:parameters_score_1} and Figure~\ref{fig:parameters_score_2} show the coherence score with different hyperparameter values on each step. 
Figure~\ref{fig:embedding_types_results} summarizes the ranges of coherence scores of each embedding model.
The experiment result shows the highest coherence score of 0.73 by using OpenAI Small embedding, with $n\_neighbors$ being 25, K-means where $n\_clusters$ being 70, $min\_df$ being 1, and vectorization happening before the training. 

\begin{table}[h]
\centering
\caption{Representative Categories Addressed in GenAI Policies}
\label{tab:genai-policies}

\begin{tabular}{ll}
\hline
\textbf{Category}                                                                                                                             & \textbf{Labels}                                                                                        \\ \hline
\begin{tabular}[c]{@{}l@{}}GenAI in \\ Class time/Learning\end{tabular}                                                                       & \begin{tabular}[c]{@{}l@{}}Allowed\\ Not Allowed\\ Not Mentioned\end{tabular}                          \\ \hline
GenAI in  Assignments                                                                                                                         & \begin{tabular}[c]{@{}l@{}}Allowed\\ Not Allowed\\ Not Mentioned\end{tabular}                          \\ \hline
GenAI in  Assessments                                                                                                                         & \begin{tabular}[c]{@{}l@{}}Allowed\\ Not Allowed\\ Not Mentioned\end{tabular}                          \\ \hline
GenAI in  Research                                                                                                                            & \begin{tabular}[c]{@{}l@{}}Allowed\\ Not Allowed\\ Not Mentioned\end{tabular}                          \\ \hline
\begin{tabular}[c]{@{}l@{}}Citing AI-Generated\\ Work\end{tabular}                                                                            & \begin{tabular}[c]{@{}l@{}}Required\\ Not Required\\ Not Mentioned\end{tabular}                        \\ \hline
\begin{tabular}[c]{@{}l@{}}Validation Requirement on \\ Information Accuracy/\\ Uses of Copyrighted Materials \\ in the Generated Output by GenAI\end{tabular} & \begin{tabular}[c]{@{}l@{}}Addressed\\ Not Addressed\end{tabular}                        \\ \hline
\begin{tabular}[c]{@{}l@{}}Cautions on \\ Information Release\end{tabular}                                                                    & \begin{tabular}[c]{@{}l@{}}Addressed\\ Not Addressed\end{tabular}                                      \\ \hline
\begin{tabular}[c]{@{}l@{}}Main Authority \\ of (Dis-)Allowing GenAI\end{tabular}                                                             & \begin{tabular}[c]{@{}l@{}}University\\ College\\ Department\\ Instructor\\ Not Mentioned\end{tabular} \\ \hline
\end{tabular}%

\end{table}

With the collected 212 documents, the authors applied the topic model with the highest coherence score and were able to discover eight categories. 
While some identified categories may have a straightforward binary decision (i.e., `allowed' or `not allowed'), some had non-binary options (Table~\ref{tab:genai-policies}). 

Many individual paragraphs or sentences were complex and ambiguous, blending various categories and proposing decisions to be made by a college, department, or instructor. Furthermore, numerous policies included templates containing a range of binary and non-binary options for instructors to apply.
Therefore, comparing the model outcomes with labels created by domain experts were used to choose the model for category classification in the next step. 

\begin{displayquote}
\label{quote:washington}
`` These tools can be inaccurate: Each individual is responsible for any content that is produced or published containing AI-generated material. Note that AI tools sometimes ``hallucinate," generating content that can be highly convincing, but inaccurate, misleading, or entirely fabricated. Furthermore, it may contain copyrighted material. It is imperative that all AI-generated content be reviewed carefully for correctness before submission or publication. It is the user’s responsibility to verify everything. "

- Washington University in St.Louis \cite{washington_b}
\end{displayquote}

\begin{displayquote}
`` Consequences of using generative AI without faculty permission:  
The use of generative AI without faculty permission will be considered a violation of the CBS Honor Code. Suspected violations of this nature will be reported to Student Conduct in the Center for Student Success and Intervention (CSSI). 
The use of generative Artificial Intelligence (AI) tools to complete an assignment or exam is prohibited unless students have a written statement from the course instructor granting permission. Unauthorized use of AI shall be treated similarly to unauthorized assistance and/or plagiarism and is subject to Dean’s Discipline. "

- Columbia Business School \cite{columbia_x}
\end{displayquote}

\begin{displayquote}
`` Option 2: Generative AI Tools Allowed—With WCP Director Consultation And Approval
\begin{itemize}
    \item Allows students to use generative AI tools as they see fit, given that they do so responsibly, transparently, and with appropriate documentation
    \item Provides students with a critical framework and set of expectations for engaging AI tools in the course and in their writing/communication processes
    \item Requires substantive attention to teaching/learning responsible and critical use of generative AI tools (i.e., giving students more freedom with these tools means providing more guidance about how to appropriately use them) "
\end{itemize}

- Georgia Tech \cite{georgia_tech_g}
\end{displayquote}

\subsubsection{Category Classification}
The category classification step identifies if there exist any categories in Table~\ref{tab:genai-policies} alongside the corresponding values from the GenAI policy statements. Due to the ambiguity of certain statements, the authors sought expert reviews and randomly selected 72 statements that were unanimously agreed on. Table~\ref{tab:policy_categories_counts} shows the categories and values assigned and used as a ground truth.

\begin{table}[h]
    \centering
    \begin{tabular}{l|l|r}
    \hline
    \textbf{Policy Category} & \textbf{Value}       & \textbf{Count} \\ \hline
    \multirow{3}{*}{Learning Use}       & Not Mentioned & 70    \\ 
                                        & Not Allowed   & 1     \\ 
                                        & Allowed       & 1     \\ \hline
    \multirow{3}{*}{Assignment Use}     & Not Mentioned & 61    \\ 
                                        & Not Allowed   & 5     \\ 
                                        & Allowed       & 6     \\ \hline
    \multirow{3}{*}{Assessment Use}     & Not Mentioned & 70    \\ 
                                        & Not Allowed   & 2     \\ 
                                        & Allowed       & 0     \\ \hline
    \multirow{3}{*}{Research Use}       & Not Mentioned & 66    \\ 
                                        & Not Allowed   & 3     \\ 
                                        & Allowed       & 3     \\ \hline
    \multirow{3}{*}{Citation Requirement} & Not Mentioned & 51   \\ 
                                        & Not Required  & 2     \\ 
                                        & Required      & 19    \\ \hline
    \multirow{2}{*}{Validation Requirement} & Not Addressed & 63  \\
                                        & Addressed     & 9     \\ \hline
    \multirow{2}{*}{Cautions on Information Release} & Not Addressed & 60 \\ 
                                        & Addressed     & 12    \\ \hline
    \multirow{5}{*}{Main Authority}    & Not Mentioned & 53    \\ 
                                        & Instructor    & 19    \\ 
                                        & Department    & 0     \\ 
                                        & College/School & 0    \\ 
                                        & University    & 0     \\ \hline
    \end{tabular}
    \caption{Policy Categories and Counts}
    \label{tab:policy_categories_counts}
\end{table} 

\begin{figure}[b]
    \centering
    \begin{subfigure}[]{\textwidth}
        \includegraphics[width=\linewidth]{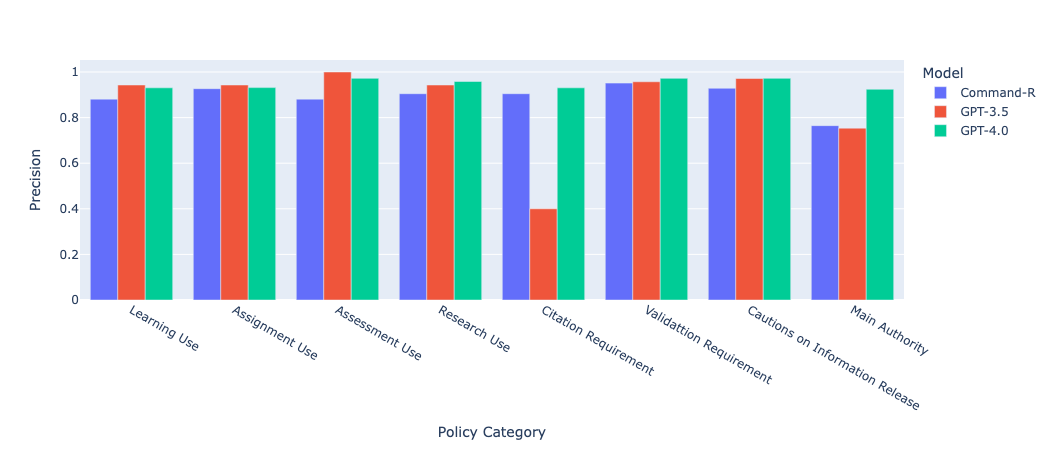}
        \caption{Precision of Classification Results}
    \end{subfigure}
    \begin{subfigure}[]{\textwidth}
        \includegraphics[width=\linewidth]{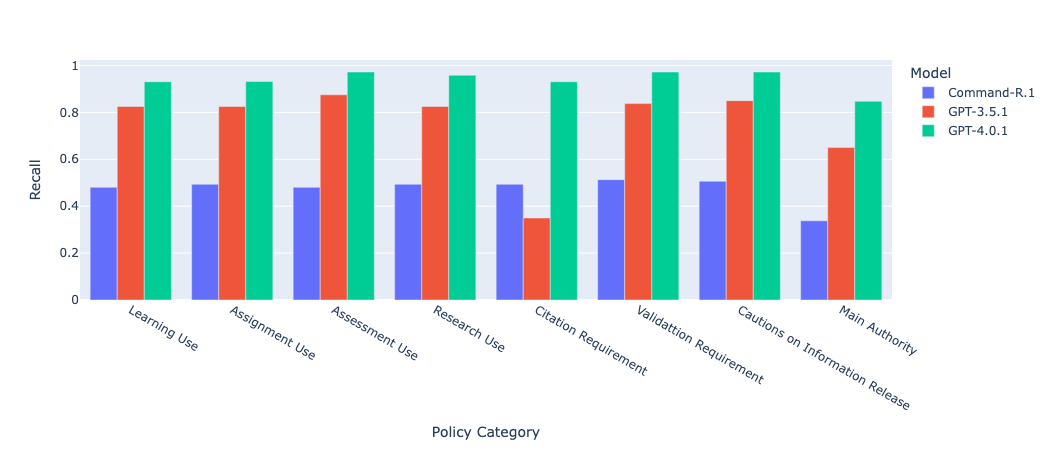}
        \caption{Recall of Classification Results}
    \end{subfigure}
    \caption{Precision and Recall of Classification Models}
    \label{fig:precision-recall}
    
\end{figure}

In this experiment, the authors applied OpenAI's GPT-3.5 and GPT-4.0 and Cohere's Command-R \cite{command_r_2024} in combination with the LangChain framework. Figure~\ref{fig:precision-recall} presents the classification results, demonstrating that GPT-4.0 outperformed the other two models in terms of both precision and recall. GPT-4.0 achieved precisions between 0.92 and 0.97, and recalls between 0.85 and 0.97. However, it is important to note that LLM performance may fluctuate over time due to factors such as outdated training data, evolving real-world contexts, and dependency on external frameworks like LangChain.

As LLM performance can vary daily, continuous monitoring is essential, and models should be re-evaluated or updated if performance falls below acceptable thresholds. Furthermore, exploring and incorporating newer or more powerful models within resource and budget constraints should be part of an ongoing evaluation strategy.

\subsection{Integration to a Web Application}

\begin{figure}[h!]
    \centering
    \begin{subfigure}[t]{\textwidth}
        \centering
         \includegraphics[width=0.4\linewidth]{Figures/PolicyEntry.png}
        \caption{AI Policy Statement Entry: In this example, the policy indicates that students may use GenAI tools for learning, but not for assignments and assessments. It also addresses that students should contact the instructor for any questions.}
        \label{fig:policy_entry}
    \end{subfigure}%
    
    \begin{subfigure}[t]{\textwidth}
        \centering
        \includegraphics[width=0.5\linewidth]{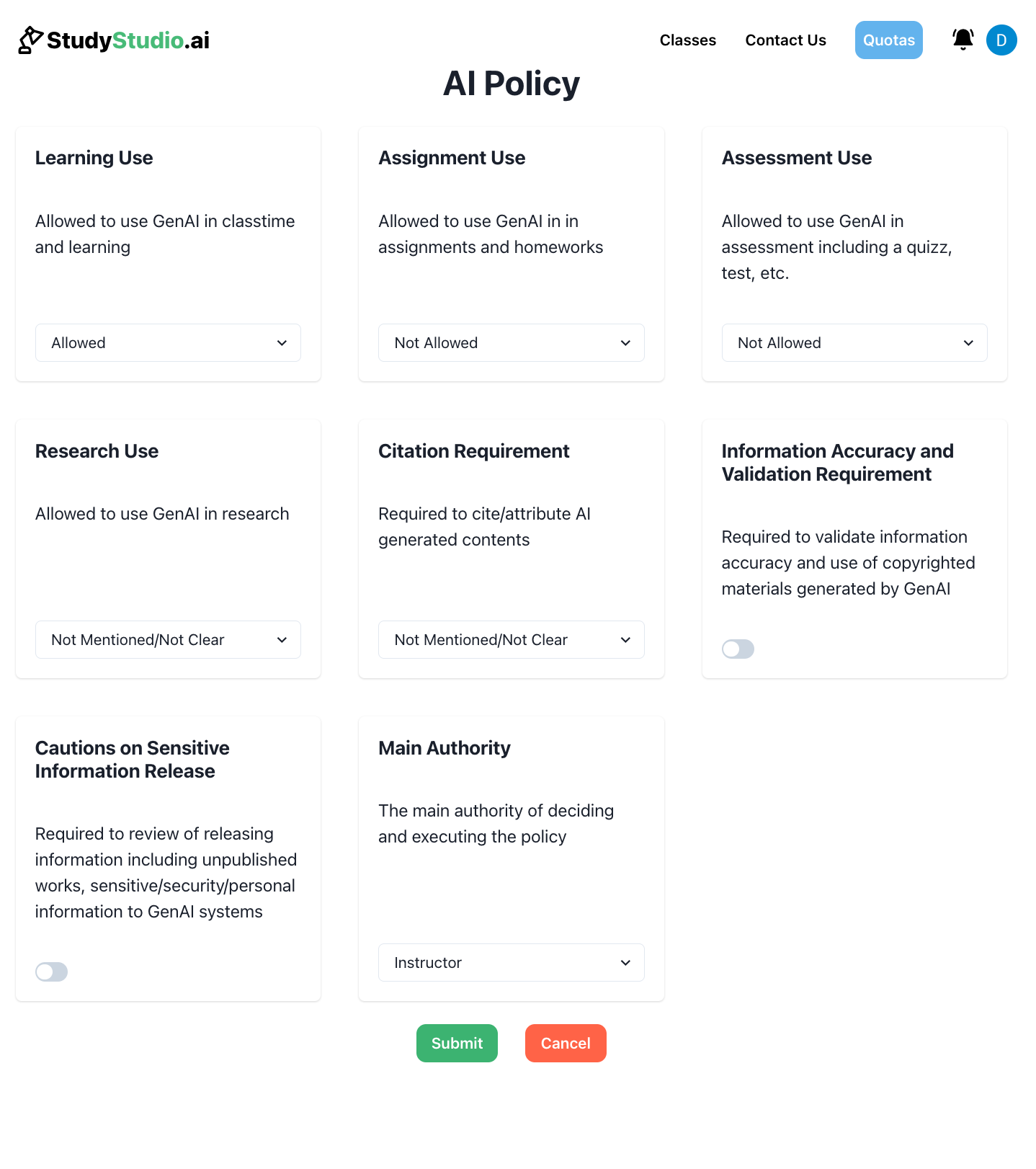}
        \caption{AI Policy Classifier/Moderator: Based on the input provided in Figure ~\ref{fig:policy_entry}, the system classifies the input across eight topics, each containing 2-5 categories. For the given input, it correctly classified learning use (allowed), assignment use (not allowed), assessment use (not allowed), and main authority (instructor), allowing the user to further adjust settings as needed.}
    \label{fig:policy_clssification}
    \end{subfigure}
    \caption{Web Interface of AI Policy Statement Entry and Classified Output}
\end{figure}

The policy topic discovery and topic classification pipeline can be applied to features on educational technologies that students may use GenAI features. For instance, StudyStudio.ai \cite{studystudio} allows to create a class or synchronize an existing class from learning management systems (LMSs), and offer features including study guides, tutor chatbot, and practice quiz generator.

Additionally, the system allows users to either upload their syllabus containing GenAI policy statements or manually enter the statements as an input (Figure~\ref{fig:policy_entry}). Submitting this information triggers the topic classification engine, which classifies the input into categories across eight topics from Section~\ref{sec:topic_discovery}. Once classified, the user can further adjust the settings, especially for categories not explicitly mentioned in the policy (Figure~\ref{fig:policy_clssification}).  For example, if a policy does not address citation requirements, the user can set this to 'required' to ensure proper attribution of sources.

Once the policy setting is confirmed, the system will limit its features or responses to comply policies. For instance, if the class doesn't allow to use AI for assignments, but the user asks a question similar to any assignment questions, the system will not provide answers and rather provide references where the user can review to start assignments on their own.

    

\section{Discussion}
\label{sec:discussion}
While GenAI can benefit student learning by providing personalized experiences and real-time feedback, significant concerns remain about how to integrate it into the current education system in alignment with pedagogical practices and intended learning outcomes. Many higher education institutions have formed committees to investigate the technology, its applications, and potential implications in order to establish policies and guidances. However, policy development varies across institutions, disciplines, and individual instructors, and under-resourced organizations may struggle to dedicate the necessary time and expertise to these efforts. Additionally, inconsistent and ambiguous policies may lead to confusion among students.

To address this challenge, the authors developed and tested a data pipeline designed to identify common topics in GenAI-related policy statements and classify categorical values from input texts.

To better understand GenAI policies in education and support the creation and evaluation of such policies, the authors collected policy statements from higher education institutions and applied topic discovery methods. This approach leveraged pre-trained word embedding models in conjunction with dimensionality reduction, clustering, vectorization, and c-TF-IDF algorithms. The output topics, selected based on high coherence scores, were reviewed by domain experts, resulting in the identification of eight frequently referenced policy themes.

Given the variability in policies across disciplines, courses, and assignments, which can lead to misinterpretation, the authors utilized large language models (LLMs) to classify values across the eight identified categories from input documents such as syllabus statements. Experimental results showed that GPT-4.0 achieved higher precision and recall compared to GPT-3.5 and Command-R. This classification capability has been incorporated into the developed application to dynamically adjust functionality, ensuring that users remain in compliance with policies while also enhancing the learning experience and achieving outcomes defined by organizations and instructors.

In future work, the authors plan to continually monitor and incorporate newly published GenAI-related policies to expand the application’s features, including an automated policy generation tool. As higher education policies continue to evolve and mature, their findings should be extended and shared with K–12 education communities as well. Since K–12 serves as a foundational stage for developing critical thinking, problem-solving, communication, collaboration, and ethical reasoning, it is essential to establish clear guidelines and policies for integrating GenAI technologies into the education of future generations \cite{adams2023ethical}.

\bibliography{references}

\end{document}